\documentclass{article}
\usepackage{spconf,amsmath,graphicx}

\usepackage{enumitem}
\setlist{nosep, leftmargin=14pt}
\usepackage{booktabs}
\usepackage{multirow}
\usepackage{mwe} 
\usepackage{amssymb}

\title{Uni-Loc: Weakly Supervised 3D CT Pathology Localization via Spatially Grounded Vision-Language Classification}
\name{Hao-Chih Lee$^\dagger$, Zahi A. Fayad, Xueyan Mei\thanks{$^\dagger$Hao-Chih.Lee@mssm.edu}}
\address{BioMedical Engineering and Imaging Institute \\ Icahn School of Medicine at Mount Sinai\\ New York, NY, USA}

\begin{document}
%
\maketitle
\begin{abstract}
Accurate localization of abnormalities in three-dimensional (3D) CT scans is essential for linking automated predictions with clinically relevant pathological findings. We propose Uni-Loc, a vision-language framework that unifies pathology classification and 3D localization using scan-level supervision. Uni-Loc derives both predictions and localization maps from spatial similarities between pathology-conditioned queries and image features. A token-to-voxel upsampling module enables high-resolution localization, while gradient-guided thresholding identifies pathology regions without additional learned parameters. Evaluated on five localization tasks comprising 355 annotated CT scans, Uni-Loc achieves macro-averaged voxel-level AUROC/AUPR of 96.1\%/34.5\%, outperforming relevance propagation (80.7\%/13.9\%) and gradient saliency (88.7\%/8.1\%). Uni-Loc also achieves classification AUROC of 88.39\% on CT-RATE and 78.03\% on RAD-ChestCT, improving over a matched baseline. These results demonstrate effective 3D pathology localization without requiring pathology-specific voxel-level supervision.
\end{abstract}
\begin{keywords}
Vision-language model, computed tomography, weakly supervised, pathology localization.
\end{keywords}
\section{Introduction}
Deep learning models have demonstrated considerable potential for automated pathology detection in 3D computed tomography (CT). However, scan-level predictions offer limited insight into the anatomical findings that drive model decisions, potentially hindering clinical interpretation and trust. Pathology-specific 3D localization maps can help radiologists inspect suspicious regions and assess whether model predictions correspond to clinically relevant abnormalities.

Learning accurate 3D pathology localization remains challenging because densely annotated medical images are scarce. Voxel-level annotations require substantial expert effort, particularly for volumetric CT scans containing hundreds of slices and abnormalities with complex boundaries. Although post-hoc explainability methods offer a convenient way to visualize model predictions, recent studies have raised concerns about their reliability in medical imaging \cite{Arun2021SaliencyTrustworthiness,Zhang2024SaliencyTrustworthiness}. Weakly supervised localization offers an alternative by learning pathology-related spatial representations from scan-level diagnostic labels, reducing the need for costly pathology-specific voxel annotations. Although weakly supervised localization has been explored in CT imaging, accurately localizing diverse pathologies in 3D volumes using only scan-level supervision remains challenging \cite{Djahnine2024WeaklySupervisedCT,Li2026PatchChestCT}.

Recently, vision-language models have shown promise in CT pathology classification
by conditioning predictions on text queries~\cite{Lee2025Uniferum}. However, BERT-style classification architectures typically produce scan-level predictions from global representations via a separate classification head, without explicitly enforcing alignment between the predicted pathology and spatial features. Consequently, models' predictions are not necessarily grounded in
spatial evidence. Weakly supervised localization addresses this by pooling instance-level scores into a bag-level prediction, but existing pooling methods assume a fixed label set ~\cite{pinheiro2015image,ilse2018attention,bai2025chestomdl}. In vision-language models, pooling of local image--text similarities has so far been explored in dual-encoder architectures, for zero-shot grounding in chest radiographs and CT~\cite{park2025radzero,park2025radzero3d}. To our knowledge, such similarity pooling has not been used in BERT-style fusion models to couple scan-level classification with dense 3D localization.

In this work, we propose Uni-Loc, a vision-language framework that unifies pathology classification and 3D localization through shared spatial similarity scores. Instead of a separate classification head predicting scan-level labels, we derive both localization maps and classification predictions from the correspondence between a pathology-conditioned query and spatial image features. We also introduce a token-to-voxel upsampling module for high-resolution localization and a gradient-guided thresholding strategy to identify pathology regions. We evaluate the framework on five pathology localization tasks using expert annotations, alongside classification benchmarks on CT-RATE and RAD-ChestCT. 

\section{Method}
\noindent {\bf Query-conditioned similarity pooling.} Our framework builds upon Uniferum, a vision-language model for query-conditioned 3D CT pathology classification \cite{Lee2025Uniferum}. Uniferum follows a BERT-style multimodal classification architecture, where visual embeddings from a CT volume and textual embeddings from a pathology-specific query are jointly processed with a learnable classification token ([CLS]) through transformer layers. The transformed [CLS] embedding is then passed through a linear classification head to produce a query-conditioned scan-level prediction. However, scan-level predictions do not directly derive from pathology-associated spatial features.

To couple classification and localization, we replace the conventional classification head with a similarity-based formulation. Specifically, the transformed [CLS] embedding instead acts as a pathology-specific query to compare with spatial image features in a shared embedding space. The resulting spatial scores provide localization signals and are aggregated through attention-weighted pooling to produce the scan-level classification logit. 

Specifically, let \(q,k_i\in\mathbb{R}^d\) denote the query embedding derived from the transformed [CLS] embedding and the vision embedding at location \(i\), respectively. We compute the localization scores \(s_i\) and classification logit \(s_{\mathrm{cls}}\) as
\begin{equation}
s_i=c\,\operatorname{sim}(q, k_i)-b,\qquad
s_{\mathrm{cls}}=\sum\nolimits_{i=1}^{N} a_i s_i,
\label{eq:att_pool}
\end{equation}
where \(\operatorname{sim}(\cdot,\cdot)\) denotes the \(\ell_2\)-normalized inner product, $b$ and $c$ are constants to scale the dynamic range of logits, and the normalized weight \(a_i\) is obtained by applying the softmax over $N$ scores \(s_i\). The resulting spatial logits $s_i$ can be reshaped into a 3D localization map, with probabilities obtained by applying a sigmoid activation if dense annotation is available. The scan-level probability is then obtained as $\sigma(s_{\mathrm{cls}})$, and the model is optimized using binary cross-entropy loss with scan-level pathology labels. We set \(c=10\) to expand the range of cosine similarities and \(b=\ln19\) to assign uncorrelated pairs (\(\operatorname{sim}\approx 0\)) a probability of about \(\sigma(-\ln 19)= 0.05\).

\noindent {\bf Gradient-guided thresholding.} We further characterize how each spatial location contributes to the classification logit:
\[
\frac{\partial s_{\textup{cls}}}{\partial s_i}
= a_i(1+s_i-s_{\mathrm{cls}}).
\]
This gradient quantifies the sensitivity with respect to each similarity score. Notably, the gradient is positive and increases monotonically with $s_i$ satisfying \(s_i>s_{\mathrm{cls}} - 1\). Such locations therefore make increasingly positive contributions to the classification logit, providing a gradient-based justification for using $s_i$ as localization signals. Accordingly, we threshold localization maps at $s_i > s_{\mathrm{cls}}-1$, which we term gradient-guided thresholding (GradT). 

\noindent {\bf Token-to-Voxel upsampling.} To overcome the 32-fold reduction in spatial resolution of Uniferum’s token embeddings and recover voxel-level localization maps, we employ a UNet-style progressive upsampler. The vision token embeddings are first projected to low-dimensional (C=16) channels to reduce memory consumption. The resulting features are progressively upsampled through five stages, each consisting of a \(3\times3\times3\) convolution followed by 3D voxel shuffle, doubling the spatial resolution along each axis. GELU activations are applied between stages, and lateral connections incorporate intermediate encoder features from the vision encoder at matching resolutions. The upsampler produces a full-resolution 16-dimensional feature map through 32-fold up-sampling that allows query-conditioned localization through equation \ref{eq:att_pool} at voxel resolution. 

\noindent {\bf Implementation details.}
\label{sec:training}
We largely follow the training protocol of Uniferum \cite{Lee2025Uniferum}, with six modifications: (1) increasing the input CT resolution from 2 mm to 1 mm isotropic spacing; (2) replacing PubMedBERT and EfficientNet-B0 with Clinical ModernBERT \cite{Lee2025ClinicalModernBERT} and EfficientNetV2-S \cite{Tan2021EfficientNetV2}, respectively; (3) pretraining the model on organ segmentation using only abdominal CT images from \cite{Blankemeier2026Merlin}; (4) incorporating standardized findings extracted from CT-RATE radiology reports as ancillary tasks, following \cite{Lai2024CARZero}; (5) fine-tuning with only the CT-RATE dataset for 100,000 iterations using a batch size of 64; and (6) maintaining an exponential moving average of model weights, updated every 10 steps with a decay factor of 0.99 for inference. All other training configurations remain consistent with the original implementation. All Uniferum-family models in this study are trained using the same recipe. All CT volumes are centered by lung segmentation and standardized to an image size of \(384 \times 320 \times 256\), with intensities clipped to \([-1000,1000]\) HU as in \cite{Lee2025Uniferum}. 

We denote the Uniferum model augmented with our U-Net upsampler for organ segmentation pretraining as Uni-CLS, which retains a linear classifier operating on the transformed [CLS] embedding. In contrast, our proposed variant, which derives scan-level predictions from localization maps using Eq.~\ref{eq:att_pool}, is denoted Uni-Loc. We note that, in Uni-CLS, the U-Net is used only for segmentation and is decoupled from classification; in Uni-Loc, the same spatial representations drive classification, segmentation, and weak localization.

\section{Experiments}
\noindent {\bf Datasets for localization benchmarks.} We evaluate weakly supervised pathology localization on five tasks from four public chest CT datasets, comprising 355 task-specific cases with voxel-level annotations. The benchmarks include five pathologies: ground-glass opacity (GGO; 10 scans) \cite{Ma2021COVID19CTSeg}, pleural effusion (PE; 78 scans) \cite{Kiser2020PleThora}, mediastinal lymphadenopathy (LAD; 120 scans) \cite{Idris2024MediastinalLymphNodeSEG}, atelectasis (AS; 35 scans) and infectious opacity (IO; 112 scans) \cite{Tsai2021RICORD}. For GGO, we exclude Radiopaedia cases without calibrated Hounsfield units. For LAD, only fully annotated studies are retained to avoid penalizing predictions on unannotated lesions. Each task is evaluated using a fixed pathology-specific prompt. We use the prompt ``Diagnosis if there is [pathology name] in the lungs.'' for Uni-CLS and Uni-Loc to generate localization maps. 

Localization performance is evaluated using voxel-level AUROC, AUPR and Dice score. AUROC and AUPR measure threshold-independent voxel-level discrimination, allowing fair comparison across methods without threshold tuning. Dice measures spatial overlap after thresholding. For consistent evaluation, all localization maps are spatially aligned to the reference CT volume and resampled to an isotropic resolution of $1\;\mathrm{mm}^3$ per voxel as in Section \ref{sec:training}.

\begin{table*}[!t]
\centering
\caption{Pathology-wise comparison of different localization methods}
\label{tab:pathology_results}
\small
\setlength{\tabcolsep}{2pt}
\renewcommand{\arraystretch}{1.}
\begin{tabular*}{\textwidth}{@{\extracolsep{\fill}}l*{10}{c}@{}}
\hline
\textbf{Method}
& \multicolumn{2}{c}{\textbf{GGO (10)}}
& \multicolumn{2}{c}{\textbf{IO (112)}}
& \multicolumn{2}{c}{\textbf{PE (78)}}
& \multicolumn{2}{c}{\textbf{LAD (120)}}
& \multicolumn{2}{c}{\textbf{AS (35)}} \\
\cline{2-3}
\cline{4-5}
\cline{6-7}
\cline{8-9}
\cline{10-11}
& AUROC & AUPR
& AUROC & AUPR
& AUROC & AUPR
& AUROC & AUPR
& AUROC & AUPR \\
\hline

Uni-Loc
& \textbf{.989} $\pm$ .01 & \textbf{.449} $\pm$ .10
& \textbf{.948} $\pm$ .07 & \textbf{.521} $\pm$ .21
& .924 $\pm$ .08 & \textbf{.441} $\pm$ .23
& \textbf{.979} $\pm$ .02 & \textbf{.218} $\pm$ .16
& \textbf{.964} $\pm$ .05 & \textbf{.098} $\pm$ .09 \\

Uni-CLS + LXT
& .885 $\pm$ .10 & .180 $\pm$ .08
& .777 $\pm$ .14 & .263 $\pm$ .14
& .839 $\pm$ .16 & .129 $\pm$ .13
& .830 $\pm$ .19 & .063 $\pm$ .07
& .703 $\pm$ .35 & .058 $\pm$ .10 \\

Uni-CLS + saliency
& .881 $\pm$ .05 & .089 $\pm$ .03
& .844 $\pm$ .05 & .152 $\pm$ .08
& .851 $\pm$ .07 & .040 $\pm$ .03
& .956 $\pm$ .02 & .091 $\pm$ .07
& .902 $\pm$ .06 & .032 $\pm$ .04 \\

C-CRM + LXT
& .756 $\pm$ .06 & .161 $\pm$ .05
& .780 $\pm$ .12 & .276 $\pm$ .16
& .815 $\pm$ .10 & .033 $\pm$ .02
& .799 $\pm$ .12 & .018 $\pm$ .03
& .584 $\pm$ .19 & .015 $\pm$ .03 \\

C-CRM (cosine)
& .870 $\pm$ .11 & .036 $\pm$ .02
& .785 $\pm$ .15 & .178 $\pm$ .19
& \textbf{.972} $\pm$ .04 & .213 $\pm$ .12
& .863 $\pm$ .13 & .048 $\pm$ .07
& .324 $\pm$ .28 & .001 $\pm$ .00 \\
\hline
\end{tabular*}

\vspace{2mm}
\parbox{\textwidth}{%
\small
Parentheses denote sample counts per pathology. Values are mean $\pm$ standard deviation; best results are bolded. C-CRM: COLIPRI-CRM
}
\end{table*}

\noindent {\bf Baseline localization methods.}
We compare our approach against conventional weak supervision, attribution-based methods and CLIP-based zero-shot localization.

For weak supervision, we evaluate a standard baseline that applies max pooling to voxel-level logits (Uni-CLS-max). As in Uni-CLS, the transformed vision embeddings are upsampled to full resolution by the U-Net upsampler and mapped to voxel-level logits, which are used for organ segmentation pretraining. During fine-tuning, the maximum over these logits serves as the scan-level logit, replacing the linear classifier on the [CLS] embedding used in Uni-CLS.

We apply layer-wise relevance propagation (LXT, \cite{achtibat2024lxt}) to Uni-CLS and COLIPRI-CRM, propagating relevance from the classification logit and image–text similarity score, respectively, to the visual tokens. Token relevance is the sum of element-wise embedding–gradient products over feature dimensions, and the resulting maps are trilinearly upsampled to input resolution. For Uni-CLS we also evaluate gradient saliency, the absolute gradient of the classification logit with respect to the input volume, which backpropagates through both the multimodal transformer and the vision encoder to yield voxel-level attributions. For CLIP-based localization, we evaluate COLIPRI-CRM \cite{Wald2026COLIPRI} in a zero-shot setting using cosine similarity between visual patch embeddings and a positive pathology prompt ``[pathology name] presents.''. The resulting $24^3$ similarity maps are trilinearly upsampled to the input resolution.

\section{Results}

\noindent {\bf Localization benchmark}. Table \ref{tab:pathology_results} compares the localization performance of Uni-Loc against baseline methods across five pathologies. Uni-Loc achieves the highest AUPR across all five pathologies and the highest AUROC in four of five, with the exception of pleural effusion, where COLIPRI (cosine) achieves a higher AUROC. Overall, Uni-Loc achieves a macro-averaged AUROC of 0.961 and AUPR of 0.345, compared with 0.807 and 0.139 for Uni-CLS + LXT and 0.887 and 0.081 for gradient saliency, respectively.

We found that Uni-CLS-max, despite achieving comparable classification performance (87.30/76.99 AUROC for CT-RATE and RAD-ChestCT, respectively), generally produces degenerate localization maps, with high logits concentrated along image borders, and achieves near-zero AUPR for all five pathologies. This highlights a limitation of max-pooling-based weak supervision, and we omit its results from Table \ref{tab:pathology_results}.

\noindent {\bf Comparison of thresholding strategies.} Table~\ref{tab:thresholding} compares GradT with a fixed threshold ($p=0.5$) and two ground-truth-tuned oracles: a global threshold ($p=0.1$) and per-pathology thresholds, selected from nine candidates (0.1--0.9). GradT achieves a macro-averaged Dice of 0.310, improving upon the fixed threshold (0.256). The improvement is driven by substantial gains in GGO (0.289 to 0.492) and IO (0.261 to 0.391), while performance decreases slightly for PE (0.408 to 0.363), LAD (0.212 to 0.208), and AS (0.109 to 0.096). GradT also slightly outperforms the global oracle (0.305), demonstrating the effectiveness of adaptive thresholding without pathology-specific threshold tuning.

\begin{table}[t]
\centering
\caption{Dice scores across thresholding strategies}
\label{tab:thresholding}
\small
\setlength{\tabcolsep}{0pt}
\begin{tabular*}{\columnwidth}{@{\extracolsep{\fill}}lcccccc@{}}
\toprule
\textbf{Threshold} & \textbf{GGO} & \textbf{PE} & \textbf{LAD} & \textbf{AS} & \textbf{IO} & \textbf{Avg.}  \\
\midrule
GradT                   & .492 & .363 & .208 & .096 & .391 & .310 \\
Fixed ($p=0.5$)          & .289 & .408 & .212 & .109 & .261 & .256 \\
Oracle global ($p=0.1$)  & \textbf{.529} & .323 & .107 & .071 & \textbf{.496} & .305 \\
Oracle per-pathology     & \textbf{.529} & \textbf{.409} & \textbf{.238} & \textbf{.117} & \textbf{.496} & \textbf{.358} \\
\bottomrule
\end{tabular*}

\vspace{3pt}
\parbox{\columnwidth}{%
\small
Oracle thresholds are selected from nine candidates
($p=0.1$--$0.9$) using ground-truth annotations.
The global oracle ($p=0.1$) maximizes macro-averaged Dice, whereas the per-pathology oracle maximizes Dice separately for each pathology (GGO: 0.1, PE: 0.6, LAD: 0.7, AS: 0.7, IO: 0.1). Best in bold.
}
\end{table}

\noindent {\bf Classification Performance Comparison.} Table \ref{tab:model_comparison} compares classification performance on CT-RATE \cite{Hamamci2026CTRATE} and the external RAD-ChestCT dataset \cite{Draelos2021RADChestCT}. Relative to our own Uni-CLS implementation, Uni-Loc improves AUPR and AUROC by 1.26 and 0.85 percentage points on CT-RATE and by 1.90 and 0.64 on RAD-ChestCT, and exceeds the fully fine-tuned Chest-OMDL in AUROC by 7.69 and 6.03 points, respectively. We include COLIPRI-CRM with linear probing as a strong CLIP-based baseline, although differences in evaluation settings limit direct comparability. Overall, the attention-based classification head enables 3D localization while slightly improving, rather than sacrificing, classification performance.

\begin{table}[t]
\centering
\caption{Comparison of classification performance.}
\label{tab:model_comparison}
\setlength{\tabcolsep}{0pt}
\small
\begin{tabular*}{\columnwidth}
{@{\extracolsep{\fill}}lcccc@{}}
\toprule
\multirow{2}{*}{\textbf{Model}} 
& \multicolumn{2}{c}{\textbf{CT-RATE}} 
& \multicolumn{2}{c}{\textbf{RAD-ChestCT}} \\
\cmidrule(lr){2-3} \cmidrule(lr){4-5}
& AUPR & AUROC & AUPR & AUROC \\
\midrule
Uni-Loc        & \textbf{65.03} & \textbf{88.39} & \textbf{55.43} & \textbf{78.03} \\
Uni-CLS        & 63.77 & 87.54 & 53.53 & 77.39 \\
Chest-OMDL \cite{bai2025chestomdl}     & --    & 80.70 & --    & 72.00 \\
COLIPRI-CRM \cite{Wald2026COLIPRI}   & 61.28 & 86.12 & 52.55 & 76.57 \\
\bottomrule
\end{tabular*}
\small
All values are reported as percentages. Best in bold.
\end{table}

\noindent {\bf Effect of  $b$.} We ablated the  offset $b$ in Eq.~\ref{eq:att_pool} by comparing $b=\ln 19$ against $b=0$ across five pathologies, using paired Wilcoxon signed-rank tests. Removing the offset significantly degraded voxel-level AUROC across all pathologies (GGO: 0.924; PE: 0.900; AS: 0.866; IO: 0.929; LAD: 0.850; all $p < 0.005$). In contrast, AUPR showed only modest changes (GGO: 0.486; PE: 0.450; AS: 0.110; IO: 0.538; LAD: 0.199), none of which were statistically significant. Classification performance remained comparable, with AUROC scores of 88.28 and 78.58 on CT-RATE and RAD-ChestCT, respectively.

\noindent {\bf Effect of attention-weighted pooling.} We compared  hard max against attention-weighted pooling in Eq.~\ref{eq:att_pool}  with fixed $b=\ln 19$. Replacing attention-weighted pooling with hard max pooling significantly degraded both voxel-level AUROC (GGO: 0.956; PE: 0.795; AS: 0.929; IO: 0.928; LAD: 0.633; all $p < 0.005$) and AUPR (GGO: 0.138; PE: 0.034; AS: 0.025; IO: 0.291; LAD: 0.003; all $p < 0.005$). Classification performance showed a slight decline, with AUROC scores of 87.66 and 77.17 on CT-RATE and RAD-ChestCT, respectively.

\begin{figure}[ht]
    \centering
    \includegraphics[width=\columnwidth]{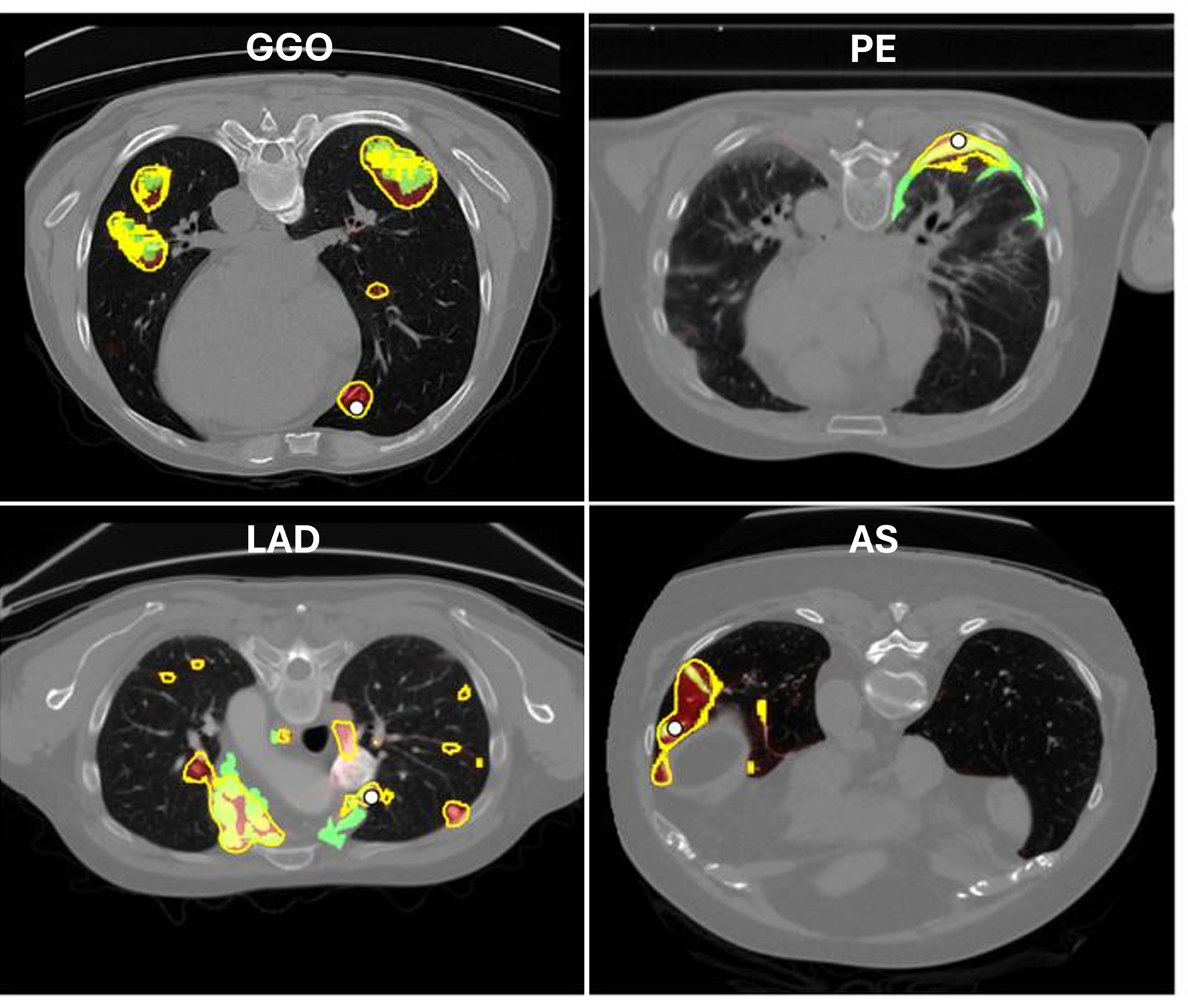} 
    \caption{Interpretable prediction with localization using Uni-Loc. Green highlights indicate expert-annotated ground-glass opacity (GGO), pleural effusions (PE), lymphadenopathy (LAD) and atelectasis (AS). Red represents the prediction probabilities of the model. The region enclosed in yellow shows the boundary outlined by the GradT rule. Samples were selected with Dice scores close to the class-wise averages in Table~2.}
    \label{fig:localization}
\end{figure}

\noindent {\bf Qualitative visualization of pathology localization}. Figure \ref{fig:localization} presents representative localization results of Uni-Loc across four chest CT abnormalities: GGO, PE, LAD, and AS. The predicted probability maps (red) show substantial spatial correspondence with expert-annotated pathology regions (green), demonstrating the model's ability to identify abnormalities with diverse spatial characteristics using scan-level supervision. For diffuse abnormalities such as GGO and AS, the model captures spatially extended pathological regions, while for PE and LAD, the responses highlight more anatomically localized structures. 

\section{conclusion}
We presented Uni-Loc, a multimodal framework that jointly performs pathology classification and 3D localization using scan-level supervision. By deriving both predictions and localization maps from shared pathology-conditioned spatial similarities, the framework explicitly grounds classification in spatial evidence. We further introduced a U-Net-based up-sampler for high-resolution localization and a parameter-free gradient-guided thresholding strategy for delineating pathology regions. Across five localization tasks comprising 355 CT scans, Uni-Loc achieved macro-averaged AUROC and AUPR of 0.961 and 0.345, outperforming relevance propagation (0.807 and 0.139) and gradient saliency (0.887 and 0.081). GradT further achieved a mean Dice of 0.310, compared with 0.256 using fixed thresholding. Uni-Loc also modestly improved classification performance on CT-RATE and RAD-ChestCT. Although localization remains challenging for certain abnormalities, these results demonstrate the potential of jointly learning classification and spatial localization without pathology-specific voxel-level annotations.

\vfill
\pagebreak
\section{Compliance with Ethical Standards}
This research study was conducted using human subject data obtained from publicly available, open-access sources. Ethical approval was not required, as confirmed by the licensing terms associated with the datasets.
\section{Acknowledgments}
This work was supported in part through the Minerva computational and data resources and staff expertise provided by Scientific Computing and Data at the Icahn School of Medicine at Mount Sinai and supported by the Clinical and Translational Science Awards (CTSA) grant UL1TR004419 from the National Center for Advancing Translational Sciences. XM is supported by the Schmidt Science Foundation. The authors used ChatGPT and Claude for language editing and grammar correction; all content was written and verified by the authors. The authors have no relevant financial or non-financial interests to disclose.
\label{sec:acknowledgments}

\label{sec:ref}
\small
\bibliographystyle{IEEEbib}
\bibliography{isbi_references}
\end{document}